\documentclass[runningheads]{llncs}

\usepackage{eccv}

\usepackage{eccvabbrv}

\usepackage{graphicx}
\usepackage{booktabs}
\usepackage{pifont}
\usepackage{capt-of}
\usepackage{subcaption}
\usepackage{multirow}
\usepackage{marvosym}

\usepackage[accsupp]{axessibility}  
\definecolor{mygray}{gray}{.5}
\definecolor{mygray2}{gray}{.9}
\newcommand{\doubleline}[2]{\begin{tabular}[c]{@{}c@{}}{#1}\\ {#2}\end{tabular}}

\usepackage{hyperref}

\usepackage{orcidlink}

\begin{document}

\title{MemPose: Category-level Object Pose Estimation with Memory} 

\titlerunning{MemPose: Category-level Object Pose Estimation with Memory}

\author{Xiao Lin\inst{1}  \and
Minghao Zhu\inst{1} \and
Yun Peng\inst{1} \and 
Liuyi Wang\inst{1} \and
Qiyi Wang\inst{1} \and \\
Chengju Liu\inst{1,2}\textsuperscript{\Letter} \and
Qijun Chen\inst{1,2}\textsuperscript{\Letter}
}

\authorrunning{Xiao Lin, et al.}

\institute{Tongji University, Shanghai, China \and
State Key Laboratory of Autonomous Intelligent Unmanned Systems \\
\email{\{linx\_xx, zmh\_hh, pengyun, wly, wqy126179, \\ liuchengju, qjchen\}@tongji.edu,cn}
}

\maketitle

\begin{abstract}
    In the pursuit of robust and generalizable category-level object pose estimation, most existing methods adopt parametric formulations that learn effective representations from data, yet they primarily encode category-level patterns into fixed shape priors or static parameter weights, which limits their scalability to highly diverse instances.
    In this paper, we rethink category-level pose estimation from a memory-centric perspective and present \textbf{MemPose}, a memory-augmented framework that explicitly incorporates category-level geometric memory into the pose estimation pipeline. 
    We introduce an external memory buffer that stores and dynamically updates structural representations from previously observed instances, enabling the model to leverage accumulated experience to support current perception.
    Extensive experiments on four challenging benchmarks (REAL275, CAMERA25, Housecat6D and Wild6D) demonstrate the superiority of our proposed method over previous state-of-the-art approaches.
\end{abstract}

\section{Introduction}
\label{sec:intro}
As a critical application in human-robot interaction~\cite{zhong2025adaptive}, the \textbf{C}ategory-level \textbf{O}bject \textbf{P}ose \textbf{E}stimation (COPE)~\cite{wang2019normalized} has attracted increasing attention. Unlike instance-level methods~\cite{lin2024transpose}, COPE is model-free and aims to estimate the 9-DoF pose for arbitrary objects within predefined categories, without relying on instance-specific CAD models. However, this setting is inherently challenging due to the significant differences among objects within the category.
To overcome these challenges, humans typically leverage memory from previous observations to perform analogical reasoning across instances.
Such memory not only supports immediate perception but is also continuously updated as new objects are encountered.

\begin{figure}[htbp]
\centering
\includegraphics[width=\columnwidth]{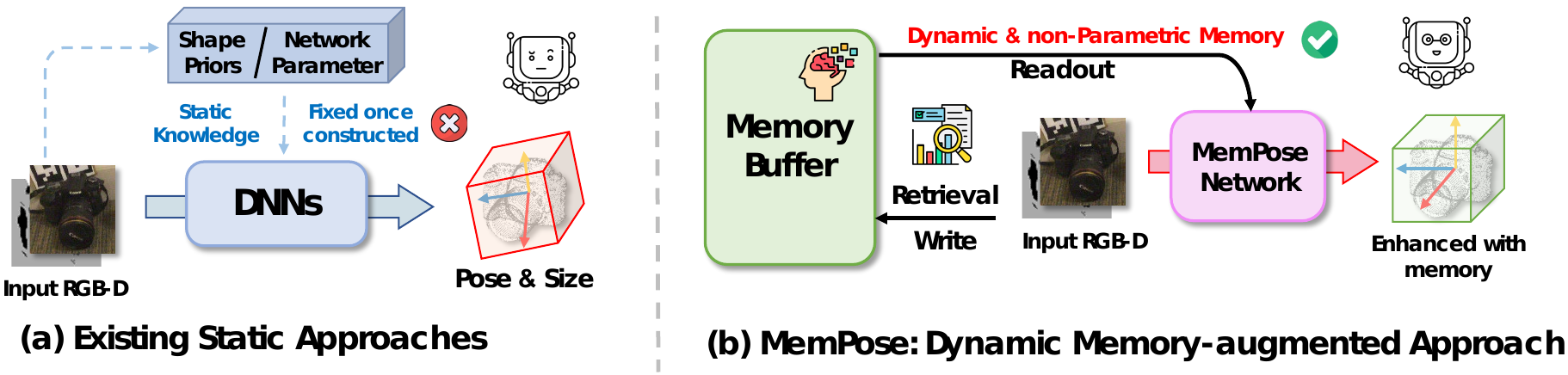}
   \caption{Overview of the category-level pose estimation pipeline: (a) Existing methods rely on static patterns, such as shape priors or fixed network parameters, to regress object pose and size. (b) In contrast, our approach introduces a dynamic, memory-augmented pipeline that explicitly incorporates category-level geometric memory to enhance pose and size estimation.
   }
   \label{fig:head_fig}
\end{figure}

Drawing on an intuitive understanding of human perception, a line of existing works~\cite{tian2020shape,chen2021sgpa,lin2022category} introduces explicit shape priors, often by extracting average shape for each category. 
Specifically, these approaches first reconstruct instance models by deforming a categorical shape prior and then match observations with the reconstructed models to regress pose. 
While priors provide explicit category-level cues, they are fixed once constructed and more like static prototypes. Essentially, such priors are hard to function as memory, as they would not update with new observations, thus capturing the diversity of instances.
Moreover, acquiring the high-quality priors requires costly pre-processing pipelines.

More recently, benefiting from the success of deep neural networks (DNNs), another line of existing methods adopt a parametric paradigm, which learns effective feature representations from input modalities via finely designed networks. 
For instance, HS-Pose~\cite{zheng2023hs} proposes a 3D graph convolution network to enhance pose-sensitive feature extraction, while AG-Pose~\cite{lin2024instance} introduces a local feature aggregation module to establish robust keypoint-level correspondences. 
KeyPose~\cite{yu2025keypose} propose a graph-based detection method to strengthen the understanding of geometric structures.
Furthermore, foundational models like DINOv2~\cite{oquab2024dinov2} have been widely adopted~\cite{chen2024secondpose,lin2025cleanpose} to enhance robustness and contextual understanding.
Although effective, these approaches implicitly encode category-level knowledge into network parameters, which remain static during inference, lacking a mechanism to provide dynamic memory support like human perception.

Despite their differences, two lines of existing methods share a common limitation: they both lack a dynamic mechanism to accumulate and reuse category-level geometric memory, as shown in \cref{fig:head_fig} (a).
This observation prompts us to rethink category-level pose estimation from a memory-centric perspective and explore how geometric memory can be effectively modeled to support robust pose prediction.

To this end, we present \textbf{MemPose}, an innovative architecture by proposing the memory module for effective and robust category-level pose estimation, as shown in \cref{fig:head_fig} (b). 
The key insight of our method is to integrate parametric perception with non-parametric memory in a unified framework. Towards this goal, we maintain a \emph{memory buffer} that stores structural feature representations. The buffer serves as an external repository of category-level knowledge, enabling the model to retrieve relevant structural patterns to support the current observation.
To ensure dynamic nature of the memory buffer, we introduce a similarity-based token merge update mechanism. As new objects are inputted, the memory buffer is continually updated with the latest keypoint context features.
Finally, by fusing memory-derived information with current features, MemPose can predict 9-DoF pose of objects.
As demonstrated by extensive experiments, our findings reveal the impact of memory-centric designs for category-level pose estimation and may inspire future work in this direction.

To summarize, our main contributions are as follows:
\begin{itemize}
    \item We introduce MemPose, a novel architecture with memory module that unifies parametric perception and non-parametric memory for category-level pose estimation.  
    \item We design a similarity-based memory update strategy to ensure the adaptability of the memory. Such s mechanism allows category-level representations to be continuously refined rather than being fixed.
    \item  Extensive experiments on three mainstream challenge benchmarks, REAL275, Housecat6D and Wild6D, demonstrate that the proposed MemPose outperforms other existing methods.
\end{itemize}

\section{Related Works}
\label{sec:related_works}
{\bf Category-level Object Pose Estimation.}
To improve the generalization ability on unseen instances, traditional methods suggest mapping input shape to a normalized canonical space (NOCS)~\cite{wang2019normalized} and recovering the pose via the Umeyama algorithm~\cite{umeyama1991least}. Furthermore, SPD~\cite{tian2020shape} proposes a deformation and matching strategy that match observations to the reconstructed models to solve poses. Inspired by SPD, many subsequent prior-based works~\cite{lin2022sar,li2025gce} further improve the use of shape priors, continuously enhancing the pose estimation performance.
More recently, prior-free methods~\cite{di2022gpv,lin2023vi,lin2024clipose,lin2024instance} have achieved impressive performance.
VI-Net~\cite{lin2023vi} separates rotation into viewpoint and in-plane rotations, while AG-Pose~\cite{lin2024instance} explicitly extracts local and global geometric keypoint information of different instances.
CleanPose~\cite{lin2025cleanpose} introduces causal learning into the formulation of COPE to mitigate the negative effects caused by confounders.
However, these methods either encode category-level structural patterns as fixed shape priors or store them within static network parameters, lacking a flexible memory mechanism to accumulate and reuse category-level experience.

\noindent
{\bf Memory-augmented Methods.}
Memory mechanism is initially introduced in Large Language Models (LLMs) to enhance long-context reasoning performance~\cite{chen2025telemem}. 
Memory mechanism has been widely used in video prediction~\cite{zhu2023fine,zhu2024mote,wang2025causalvtg}, point tracking~\cite{dong2025online} and robotic tasks~\cite{shi2025memoryvla,sheng2026dream,liu2026stamp}, demonstrating their effectiveness. For instance, Memflow~\cite{dong2024memflow} leverages memory for optical flow estimation, while MemoryVLA~\cite{shi2025memoryvla} explicitly incorporates memory to model temporal dependencies in robotic manipulation. 
In thr 3D domain, MAD~\cite{agro2025mad} constructs a memory bank to store past prediction and trajectory for 3D detection. 
\begin{figure*}[htbp]
    \includegraphics[width=\textwidth]{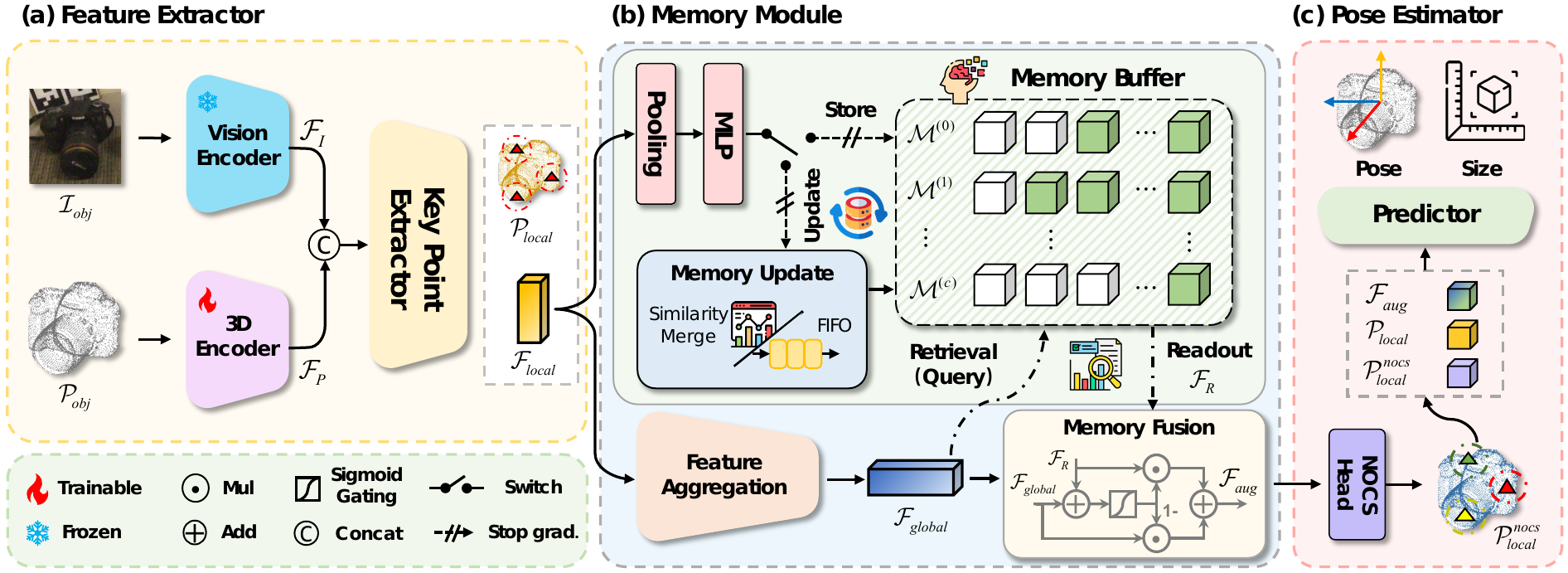}
    \caption{Illustration of MemPose. (a) The model extracts semantic and geometric features from RGB-D inputs and detects robust keypoints to obtain keypoint-level representations.
    (b) A dynamic memory module is introduced to store and retrieve historical feature representations. Through attention-based querying and memory fusion, the model leverages short-term geometric information to enhance robust feature learning.
    (c) The fused keypoint features, together with the predicted keypoints and NOCS coordinates, are utilized for joint object pose and size estimation.
    }\label{fig:MemPose}
\end{figure*}
Some approaches also attempt to utilize memory mechanism to solve point cloud tracking~\cite{xu2023mbptrack} and 3D reconstruction~\cite{wang20253d}.
However, above methods primarily focus on temporal modeling, it is non-trivial to adapt these approaches to pose estimation due to the inherent differences in human modeling among these tasks.
In this work, our MemPose exploits category-level structural patterns to form memory mechanism. Importantly, our motivation originates from a deep analysis of intra-category generalization in pose estimation, which is naturally suitable for modeling with memory.

\section{Methodology}
\label{sec:methodology}
\subsection{Definition and Overview}
{\bf Problem Setup.}
Given an RGB-D image containing objects from a predefined set of categories, we first employ segmentation masks to obtain the cropped RGB image $\mathcal{I}_{obj} \in \mathbb{R}^{H \times W \times 3}$ and the point cloud $\mathcal{P}_{obj} \in \mathbb{R}^{N \times 3}$, where $N$ is the number of points and $\mathcal{P}_{obj}$ is acquired by back-projecting the cropped depth image with camera intrinsics followed by a downsampling process. With the input $\mathcal{I}_{obj}$ and $\mathcal{P}_{obj}$, the objective of COPE~\cite{wang2019normalized} is to recover the 9-DoF pose of the target object, including the 3D rotation $R \in SO(3)$, the 3D translation $t \in \mathbb{R}^3$, and size $s \in \mathbb{R}^3$.

\noindent
{\bf Overview.} 
We present an overview of MemPose as in \cref{fig:MemPose}. Specifically, our method first encodes the current observation to construct an RGB-D representation. To explicitly capture geometric information, we further perform keypoint detection to extract local features (\cref{sec:feature_extraction}). Then, these local structural patterns are stored in a memory buffer $\mathcal{M}$, forming a short-term memory. By feature enqueue operation, memory buffer is continually updated with the latest geometric context (\cref{sec:memory_module}). Subsequently, the aggregated global feature reads from the memory and fuses the retrieved information. Finally, MemPose regresses the object pose based on the fused feature representation (\cref{sec:estimation_and_loss}).

\subsection{Partial Feature Extraction}
\label{sec:feature_extraction}
Following~\cite{lin2024instance,lin2025cleanpose}, we utilize the PointNet++~\cite{qi2017pointnet++} to extract point feature $\mathcal{F}_{P} \in \mathbb{R}^{N \times C_1}$ of input point cloud $\mathcal{P}_{obj} \in \mathbb{R}^{N \times 3}$.
As for the RGB image $\mathcal{I}_{obj}$, we adopt DINOv2 (ViT-S/14)~\cite{oquab2024dinov2} as our image feature extractor, which has been proven to extract abundant semantic-aware information from RGB images~\cite{peng2025sam,peng2026vllm}. We select those pixel features corresponding to $\mathcal{P}_{obj}$ and utilize linear interpolation to propagate the original DINOv2 features into the final RGB features $\mathcal{F}_{I} \in \mathbb{R}^{N \times C_2}$. Moreover, we concatenate $\mathcal{F}_{P}$ and $\mathcal{F}_{I}$ to form $\mathcal{F}_{obj} \in \mathbb{R}^{N \times C}$.
The local geometric information is indispensable to establish robust correspondences. Typically, methods like Farthest Point Sampling (FPS) can be used to extract local keypoints. In this work, we follow previous method~\cite{lin2024instance} and utilize a instance-adaptive approach that focuses on the most discriminative and reliable object regions.
Similar to DETR~\cite{carion2020end}, we initial $M$ learnable category embeddings $\mathcal{F}_{cat} \in \mathbb{R}^{M \times C}$, which undergoes cross-attention with $\mathcal{F}_{obj}$ to attend to critical regions in $\mathcal{P}_{obj}$. This process obtains an instance feature $\mathcal{F}_{ins} =$ CrossAttn($\mathcal{F}_{cat}, \mathcal{F}_{obj}$), We then compute correspondences between $\mathcal{F}_{ins}$ and $\mathcal{F}_{obj}$ via cosine similarity, forming a matrix $H \in \mathbb{R}^{M \times N}$, and $M$ keypoints are selected as $\mathcal{P}_{local} = \text{softmax}(H)\mathcal{P}_{obj}$, along with their corresponding feature $\mathcal{F}_{local} = \text{softmax}(H)\mathcal{F}_{obj}$.

\subsection{Memory Module}
\label{sec:memory_module}
To better exploit short-term working memory, we propose a novel memory module and processing pipeline for the COPE task. Specifically, we maintain a category-aware memory buffer defined as follows:
\begin{small}
\begin{equation}
\label{equ:memory_buffer}
    \mathcal{M} = \{\mathcal{M}^{(c)} \mid c = 0,\dots,k \},     \mathcal{M}^{(c)} \in \mathbb{R}^{L \times C}
\end{equation}
\end{small}
where $k$ denotes the number of object categories, and $\mathcal{M}^{(c)}$ stores the memory features associated with category $c$. Here, $L$ represents the length of memory buffer.

\noindent
{\bf Memory Construction.}
Each memory buffer $\mathcal{M}^{(c)}$ is initialized as an empty set. During training, newly batch features are first pushed into the corresponding category buffer. The memory retrieval and update mechanisms are activated only after the buffer reaches its capacity $L$. This warm-up strategy effectively prevents noisy and unstable representations of the early training stage, improving the robustness of the stored patterns.
Note that, most existing approaches~\cite{dong2024memflow,lin2025cleanpose} directly store the final output features as memory entries. However, such a design typically affects the attention-based retrieval process by biasing it towards recently updated entries, as these features are more closely aligned with the current network state rather than being geometrically relevant. As a result, memory retrieval may be dominated by temporal recency instead of meaningful structural similarity.
To this end, our design stores pooled local features as memory entries. Specifically, we first employ a simple average pooling on local features and get $\mathcal{F}^{avg}_{local} \in \mathbb{R}^{1 \times C}$. Then, we linear project pooled features to form memory entries $\mathcal{F}_{mem}$,
\begin{small}
\begin{equation}
\label{equ:query1}
    \mathcal{F}^{avg}_{local} = \text{AvgPool} (\mathcal{F}_{local})
\end{equation}
\begin{equation}
\label{equ:query2}
    \mathcal{F}_{mem} = \text{Norm}(\text{MLP}(\mathcal{F}^{avg}_{local})), 
\end{equation}
\end{small}
where the pooled local features capture explicit local geometric patterns, providing a more stable and geometry-aware reference for subsequent memory retrieval.

\noindent
{\bf Update Mechanism.}
The memory update process is performed before memory retrieval to ensure that the stored representations reflect the most informative geometric context.
When the number of stored entries exceeds $L$, the update mechanism is activated. Specifically, given a newly incoming pooled local feature $\mathcal{F}_{mem}$, we compute its cosine similarity with all existing memory entries in the corresponding buffer $\mathcal{M}^{(c)}$. The memory entry with the highest similarity is selected and updated by averaging it with the incoming feature. 
\begin{small}
\begin{equation}
\label{equ:argmax_base}
i^{*} = \arg\max_{i} \; \cos\!\left(\mathcal{F}_{mem}, \mathcal{M}^{(c)}_i\right),
\quad i = 1, \dots, l,
\end{equation}
\begin{equation}
\label{equ:update_base}
\mathcal{M}^{(c)}_{i^{*}} 
\leftarrow 
\frac{1}{2}
\left(
\mathcal{M}^{(c)}_{i^{*}} 
+
\mathcal{F}_{mem}
\right).
\end{equation}
\end{small}
This update mechanism mitigates memory bloat by reducing redundancy.
Furthermore, to prevent imbalance in similarity-based feature selection, we introduce several lightweight balancing strategies. For instance, the update target is randomly sampled from the top-$k$ most similar memory entries, or mild stochastic perturbations are applied during the update process. 
Here, we adopt Gumbel noise $g$~\cite{jang2017categorical} to add random perturbations
\begin{equation}
    g = -\log(-\log(u))
\end{equation}
\begin{equation}
    \cos\!\left(\mathcal{F}_{mem}, \mathcal{M}^{(c)}_i\right) \leftarrow \cos\!\left(\mathcal{F}_{mem}, \mathcal{M}^{(c)}_i\right)+\lambda g,
\end{equation}
where $u \sim \mathcal{U}(0, 1)$ is drawn from a uniform distribution, and $\lambda$ (set to 0.1) is a scaling factor controlling the magnitude of the perturbation.
Meanwhile, the classic FIFO strategy can also be used to update the buffer.
The effectiveness of these strategies is validated in ablation studies in \cref{sec:ablation_studies}.

\noindent
{\bf Memory Retrieval and Fusion.}
Recall that we expect the model to focus on geometric semantics rather than temporal similarity during retrieval. To this end, we first enhance feature representations with explicit geometric context before interacting with the memory. Specifically, we apply a geometric aggregation approach~\cite{lin2024instance} on the local features to form a global feature representation $\mathcal{F}_{global} \in \mathbb{R}^{M \times C}$, which is jointly enriched by the local geometric details from $k$-nearest neighbors and the global structural information across all keypoints.
Subsequently, we treat global features as the query for memory retrieval and stack all memory entries stored in the category-specific memory buffers $\mathcal{M}^{(c)}$ to form a unified tensor, which serves as the key–value set
\begin{small}
\begin{equation}
\mathcal{F}'_{mem} = \left[\mathcal{F}_{{mem}}^{0};\cdots;\mathcal{F}_{{mem}}^{L}\right] \in \mathbb{R}^{L \times C}
\end{equation}
\begin{equation}
q = \mathcal{F}_{global}W_q, \; k = \mathcal{F}'_{mem}W_k, \; v=\mathcal{F}'_{mem}W_v,
\end{equation}
\end{small}
where $[;]$ is the concatenation operation along the first dimension, and $W_q,W_k,W_v$ are the learnable projection parameters,
Hence, the retrieved memory features can be read-out by
\begin{small}
\begin{equation}
\mathcal{F}_{R}=\text{Softmax}(1 / \sqrt{D_k}\times q \times k^T)\times v.
\end{equation}
\end{small}
Furthermore, to enhance the stability of memory learning, we introduce an adaptive gate fusion method to integrate memory retrieved features and current global features.
\begin{small}
\begin{equation}
w_{a} = \sigma(\mathcal{F}_{R}W_R + \mathcal{F}_{global}W_f)
\end{equation}
\begin{equation}
\mathcal{F}_{aug} = w_{a}\odot \mathcal{F}_{R} + (1-w_{a})\odot \mathcal{F}_{global},
\end{equation}
\end{small}
where $\sigma$ and $\odot$ mean the Sigmoid gate function and element-wise multiplication. $W_{R/f} \in \mathbb{R}^{C \times 1}$ is a learnable weight parameter. The resulting memory-augmented features $\mathcal{F}_{aug}$ is then forwarded to the pose and size predictor.

\subsection{Pose Estimation and Loss Function}
\label{sec:estimation_and_loss}
With the obtained memory-enhanced features, we follow~\cite{lin2024instance} to perform self-attention and MLP modules to predict the corresponding NOCS coordinates $\mathcal{P}^{nocs}_{local} \in \mathbb{R}^{M \times 3}$. 
Then, given the NOCS coordinates of keypoints $\mathcal{P}^{nocs}_{local}$, the memory-enhanced features $\mathcal{F}_{aug}$ and the position of keypoint $\mathcal{P}_{local}$, we recover the final pose and size $R,t,s$ via a set of keypoint-level correspondences containing global features and points. We use L2-norm Loss to supervise the predicted pose, in formula:
\begin{small} 
\begin{equation}
\label{equ:pose_rts}
    \mathcal{L}_{pose} = \left\|R_{gt} - R\right\|_{2} + \left\|t_{gt} - t\right\|_{2} + \left\|s_{gt} - s\right\|_{2},
\end{equation}
\end{small}
where $R_{gt}, t_{gt}, s_{gt}$ means the ground truth rotation, translation and size. 

In addition, there are some additional loss functions to balance keypoints selection and pose prediction. First, to encourage the keypoints to focus on different parts, the diversity loss $\mathcal{L}_{div}$ is used to force the detected keypoints to be away from each other, in detail:
\begin{small} 
\begin{align}
    \mathcal{L}_{div}&=\sum_{i=1}^{N_{local}} \sum_{j=1, j \neq i}^{N_{local}} \mathbf{d}\left(\mathcal{P}_{local}^{(i)}, \mathcal{P}_{local}^{(j)}\right) \label{equ:supp_divloss1}\\
    \mathbf{d}\left(\mathcal{P}_{local}^{(i)}, \mathcal{P}_{local}^{(j)}\right)&=\max \left\{th_{1}-\left\|\mathcal{P}_{local}^{(i)}-\mathcal{P}_{local}^{(j)}\right\|_{2}, 0\right\} \label{equ:supp_divloss2},
\end{align}
\end{small}
where $th_1$ is a hyper-parameter and is set as 0.01, $\mathcal{P}_{local}^{(i)}$ means the $i$-th keypoint.
Then, to encourage the keypoints to locate on the surface of the object and exclude outliers simultaneously, an object-aware chamfer distance loss $\mathcal{L}_{ocd}$ is employed to constrain the distribution of $\mathcal{P}_{local}$. In formula:
\begin{small} 
\begin{align}
    \mathcal{L}_{ocd}=\frac{1}{\left|\mathcal{P}_{local}\right|} \sum_{x_{i} \in \mathcal{P}_{local}} \min _{y_{j} \in \mathcal{P}_{obj}^{'}}\left\|x_{i}-y_{j}\right\|_{2}, \label{equ:supp_ocdloss}
\end{align}
\end{small}
where $\mathcal{P}_{obj}^{'}$ denotes the point cloud of objects without outlier points.
Moreover, we also use MLP to predict the NOCS coordinates of keypoints $\mathcal{P}_{local}^{nocs} \in \mathbb{R}^{N_{local} \times 3}$. Then, we generate ground truth NOCS of keypoints $\mathcal{P}_{local}^{gt}$ by projecting their coordinates under camera space $\mathcal{P}_{local}$ into NOCS using the ground truth $R_{gt},t_{gt},s_{gt}$. And we use the $SmoothL_1$ loss to supervise the NOCS projection:
\begin{small} 
\begin{align}
    \mathcal{P}_{local}^{gt}&=\frac{1}{\left\|s_{gt}\right\|_{2}} R_{gt}\left(\mathcal{P}_{local}-t_{gt}\right) \label{equ:supp_nocsgt}\\
    \mathcal{L}_{nocs} &= SmoothL_{1}(\mathcal{P}_{local}^{gt},\mathcal{P}_{local}^{nocs}). \label{equ:supp_nocsloss}
\end{align}
\end{small}
Furthermore, to ensure that our keypoints and associated features effectively represent the partial observation $\mathcal{P}_{local}$, we additionally employ a reconstruction module to recover its 3D geometry. This module takes keypoint positions and features as input, applies positional encoding to the keypoints, and refines their features through a MLP. The encoded and refined features are aggregated, and a shape decoder predicts reconstruction deltas to recover the geometry. The reconstruction loss is defined as the object-aware Chamfer distance (CD) between the partial observation $\mathcal{P}_{local}$ and the reconstructed point cloud $\mathcal{P}_{recon}$:
\begin{small} 
\begin{align}
    \mathcal{L}_{rec}=\frac{1}{\left|\mathcal{P}_{{recon }}\right|} \sum_{x \in \mathcal{P}_{ {recon}}} \min _{y \in \mathcal{P}_{local}^{*}}\|x-y\|_{2}.
\end{align}
\end{small}

Finally, the NOCS loss $\mathcal{L}_{nocs}$ is used to supervise the prediction of $\mathcal{P}^{nocs}_{local}$. ence, the overall loss function is as follow
\begin{small} 
\begin{equation}
\label{equ:overall_loss_func}
    \mathcal{L}_{all} = \alpha_{1}\mathcal{L}_{pose} + \alpha_{2}\mathcal{L}_{div}  + \alpha_{3}\mathcal{L}_{ocd}  + \alpha_{4}\mathcal{L}_{rec}  + \alpha_{5}\mathcal{L}_{nocs},
\end{equation}
\end{small}
where $\alpha_{1}, \alpha_{2}, \alpha_{3}, \alpha_{4}, \alpha_{5}$ are hyper-parameters to balance the contribution of each term.

\begin{table}[htbp]
    \scriptsize 
    \centering
    \setlength\tabcolsep{1.8pt}
    \begin{tabular}{l|c|c|ccc|cc|ccc}
    \toprule
    \multicolumn{2}{l|}{Dataset} & \multicolumn{4}{c|}{REAL275} & \multicolumn{5}{c}{Housecat6D} \\
    \midrule
     Methods & Prior & $IoU_{75}$  & 5°2\emph{cm}  & 5°5\emph{cm} & 10°2\emph{cm}   & $IoU_{50}$  & $IoU_{75}$  & 5°2\emph{cm}    & 10°2\emph{cm}  & 10°5\emph{cm} \\
    \midrule
    $\text{SGPA}_{\text{\textcolor{mygray}{\, ICCV'21}}}$~\cite{chen2021sgpa}  & \ding{51}     & 61.9         & 35.9 & 39.6    & 61.3    & - & - & - & -  & - \\
    $\text{RBP-Pose}_{\text{\textcolor{mygray}{\, ECCV'22}}}$~\cite{zhang2022rbp}   & \ding{51}    & 67.8         & 38.2 & 48.1         & 63.1   & - & - & - & -  & - \\
    $\text{DPDN}_{\text{\textcolor{mygray}{\, ECCV'22}}}$~\cite{lin2022category}  & \ding{51}     & 76.0        & 46.0         & 50.7        & 70.4     & - & - & - & - & -  \\
    $\text{GCE-Pose}_{\text{\textcolor{mygray}{\, CVPR'25}}}$~\cite{li2025gce}  & \ding{51}     & 79.8        & 57.0         & 65.1        & 75.6     & 76.1 & \underline{55.6} & \underline{22.2}  & 49.3 & 53.5  \\
    \midrule
    $\text{FS-Net}_{\text{\textcolor{mygray}{\, CVPR'21}}}$~\cite{chen2021fs}   & \ding{55}     & -        & -         & 28.2        & -    & 48.0 & 14.8 & 3.3  & 17.1 & 21.6  \\
    $\text{GPV-Pose}_{\text{\textcolor{mygray}{\, CVPR'22}}}$~\cite{di2022gpv}  & \ding{55}      & 64.4         & 32.0 & 42.9         & -   & 50.7 & 15.2 & 3.5  & 17.8 & 22.7 \\
    $\text{VI-Net}_{\text{\textcolor{mygray}{\, ICCV'23}}}$~\cite{lin2023vi}    & \ding{55}     &48.3       & 50.0          &57.6         & 70.8  & 56.4 & 20.4 & 8.4 & 20.5 & 29.1         \\
    $\text{PRD-Pose}_{\text{\textcolor{mygray}{\, ICCV'25}}}$~\cite{lee2025joint}   & \ding{55}      &-       & 52.6          &-         & 73.4  & - & - & 8.9  & 26.2 & -         \\
    $\text{SecondPose}_{\text{\textcolor{mygray}{\, CVPR'24}}}$~\cite{chen2024secondpose} & \ding{55}        &77.7       & 56.2          &63.6         & 74.7     & 66.1 & 24.9 & 11.0 & 25.3 & 35.7       \\
    $\text{AG-Pose}_{\text{\textcolor{mygray}{\, CVPR'24}}}$~\cite{lin2024instance}    & \ding{55}     &80.1       & 57.0         &64.6         & 75.1      & \underline{76.9} & 53.0 & 21.3 & \underline{51.3} & 54.3        \\
    $\text{KeyPose}_{\text{\textcolor{mygray}{\, AAAI'25}}}$~\cite{yu2025keypose}   & \ding{55}      &\underline{80.8}       & 57.7         &66.0         & \underline{78.8}      & 75.4 & - & - & -  & -        \\
    $\text{SpherePose}_{\text{\textcolor{mygray}{\, ICLR'25}}}$~\cite{ren2025learning} & \ding{55}        &79.0       & \underline{58.2}         &\underline{67.4}         & 76.2     & 72.2 & - & 19.3 & 40.9 & \underline{55.3}        \\

    \midrule
    \textbf{MemPose (ours)}   & \ding{55}    & \textbf{81.0} & \textbf{59.9}         & \textbf{67.7} & \textbf{79.0}    & \textbf{81.5} & \textbf{56.4} & \textbf{23.1}  & \textbf{52.6} & \textbf{57.3}   \\
    \bottomrule
    \end{tabular}
    \caption{\textbf{Comparison with state-of-the-art methods on REAL275 and Housecat6D dataset.}  A higher value indicates better performance. ‘-’ means unavailable statistics. Overall best results are in \textbf{bold} and the second best results are \underline{underlined}.
    }
    \label{tab:compare_sota}
\end{table}

\section{Experiments}
\label{sec:Experiments}
{\bf Datasets and Metrics.}
Following previous works, we conduct experiments on four mainstream benchmarks, REAL275, CAMERA25~\cite{wang2019normalized}, HouseCat6D~\cite{jung2024housecat6d} and Wild6D~\cite{fu2022category} datasets.
REAL275 is a challenging real-world dataset that contains objects from six categories. The training data consists of 4.3k images from 7 scenes, while the testing data includes 2.75k images from 6 scenes and 3 objects from each category. 
HouseCat6D is a comprehensive multi-modal real-world dataset and encompasses ten household categories, including photometrically challenging objects like glass and cutlery, with occlusions.
Wild6D is a large dataset designed for self-supervised learning of COPE. It is only annotated on the test set with images from 486 different background videos, containing 162 objects from five categories. In this work, we only use the test set of Wild6D for evaluation to enrich the real-world experiments.
When validating REAL275 and Wild6D, we also used CAMERA25~\cite{wang2019normalized} for training, which is a synthetic dataset that contains the same categories as REAL275. It provides 300k synthetic RGB-D images of objects rendered on virtual backgrounds, with 25k images are withheld for testing.

We evaluate the model performance with two metrics. (\textbf{i}) \textbf{3D IoU}. As for the NOCS dataset, we report the mean average precision (mAP) of Intersection over Union (IoU) with the thresholds of 75\%. For the HouseCat6D dataset, we report the mAP of 3D IoU under thresholds of 25\%, 50\% and 75\%. (\textbf{ii}) \textbf{n°m \emph{cm}}.  We also utilize the combination of rotation and translation metrics of 5°2 \emph{cm}, 5°5 \emph{cm}, 10°2 \emph{cm} and 10°5 \emph{cm}, which means the estimation is considered correct when the error is below a threshold. 

\noindent
{\bf Implementation Details.}
For a fair comparison, we utilize the same segmentation masks as AG-Pose~\cite{lin2024instance} and DPDN~\cite{lin2022category} from MaskRCNN~\cite{he2017mask} and resize them to $224 \times 224$. 
For model parameters, the number of points $N$ in point cloud is 1024 and the number of keypoints $M$ is set as 96.  the feature dimensions are set as $C_1 = C_2 =$ 128 and $C =$ 256, respectively.
For memory module, we set the size of category-specific buffer to 96. The number of object categories $k$ is 6 in datasets REAL275 and Wild6D, and 10 in dataset Housecat6D. Technically, the memory buffer is registered as a buffer tensor during training via $\mathtt{register\_buffer}$, which would be saved and restored together with the model. It does not participate in gradient backpropagation and is not updated by the optimizer.
For the hyper-parameters in the loss functions, $\alpha_{1}, \alpha_{2}, \alpha_{3}, \alpha_{4}, \alpha_{5}$ are $0.3,10.0,2.0,15.0,2.0$, respectively. 
\begin{table}[htbp]
    \scriptsize
    \centering
    \setlength\tabcolsep{6pt}
    \begin{tabular}{l|c|cc|ccccc}
    \toprule
   Methods & Prior & $IoU_{50}$ & $IoU_{75}$  & 5°2\emph{cm} & 5°5\emph{cm} & 10°2\emph{cm}  & 10°5\emph{cm} \\
    \midrule
    $\text{SPD}_{\text{\textcolor{mygray}{\, ECCV'20}}}$~\cite{tian2020shape} & \ding{51}  & 93.2    & 83.1              & 54.3      &59.0          & 73.3    &81.5     \\
    $\text{SGPA}_{\text{\textcolor{mygray}{\, ICCV'21}}}$~\cite{chen2021sgpa}   & \ding{51}                    &93.2     &88.1           &70.7        & 74.5           &82.7    &88.4    \\
    $\text{RBP-Pose}_{\text{\textcolor{mygray}{\, ECCV'22}}}$~\cite{zhang2022rbp}  & \ding{51}   & 93.1     & 89.0          & 73.5      & 79.6          & 82.1   &89.5    \\
    \midrule
    $\text{NOCS}_{\text{\textcolor{mygray}{\, CVPR'19}}}$~\cite{wang2019normalized} & \ding{55}  & 83.9    & 69.5              & 32.3      &40.9          & 48.2    &64.4     \\
    $\text{DualPoseNet}_{\text{\textcolor{mygray}{\, ICCV'21}}}$~\cite{lin2021dualposenet}      & \ding{55}        &92.4         &86.4                &64.7        & 70.7           &77.2    &84.7    \\
    $\text{GPV-Pose}_{\text{\textcolor{mygray}{\, CVPR'22}}}$~\cite{di2022gpv}   & \ding{55}    & 93.4        & 88.3            & 72.1      & 79.1          & -    & 89.0     \\
    $\text{HS-Pose}_{\text{\textcolor{mygray}{\, CVPR'23}}}$~\cite{zheng2023hs}   & \ding{55}    & 93.3        & 89.4            & 73.3      & 80.5          & 80.4    & 89.4     \\
    $\text{VI-Net}_{\text{\textcolor{mygray}{\, ICCV'23}}}$~\cite{lin2023vi} & \ding{55}   &-              &-              & 74.1     & 81.4         & 79.3  & 87.3     \\
    $\text{CLIPose}_{\text{\textcolor{mygray}{\, TCSVT'24}}}$~\cite{lin2024clipose} & \ding{55}   &-              &91.0             & 74.8     & 82.2         & 82.0    &  90.6     \\
    $\text{AG-Pose}_{\text{\textcolor{mygray}{\, CVPR'24}}}$~\cite{lin2024instance} & \ding{55}   &93.8              &91.3              & 77.8     & 82.8         & \underline{85.5}  & 91.6     \\
    $\text{SpherePose}_{\text{\textcolor{mygray}{\, ICLR'25}}}$~\cite{ren2025learning} & \ding{55}   &\textbf{94.8}              &\underline{92.4}              & \underline{78.3}     & \underline{84.3}         & 84.8  & \underline{92.3}     \\
    \midrule
    \textbf{MemPose (ours)}  & \ding{55}  &\underline{94.2}                &\textbf{92.5}             & \textbf{78.4}     & \textbf{84.6}         & \textbf{87.0}    &  \textbf{92.6}     \\
    \bottomrule
    \end{tabular}
    \caption{\textbf{Comparisons with state-of-the-art methods on CAMERA25 dataset.} A higher value indicating better performance. ‘-’ means unavailable statistics.  Overall best results are in \textbf{bold} and the second best results are \underline{underlined}.
    }
    \label{tab:compare_sota_camera}
\end{table}
For model optimizing, we employ the same data augmentation approach as previous works~\cite{lin2024instance,lin2022category}, which leverage random rotation degree sampled from $U$(0, 20) and rotation $\Delta t \sim U$(-0.02, 0.02) and scaling $\Delta s \sim U$(-0.18, 1.2).
We train the model on a single NVIDIA L40 GPU for a total of 120k iterations by the Adam~\cite{kingma2014adam} optimizer, with a mini training batch is 36 and a learning rate range from 2e-5 to 5e-4 based on triangular2 cyclical schedule~\cite{smith2017cyclical}.

\subsection{Comparison with State-of-the-Art Methods}
\label{sec:compare_sota}
{\bf Results on REAL275 and Housecat6D datasets.}
\cref{tab:compare_sota} shows the comparison of our method with previous methods on REAL275 and Housecat6D datasets. Overall, our MemPose consistently outperforms previous methods and achieves state-of-the-art performance on both datasets.
On REAL275, our method obtains the best results on all reported metrics. In particular, when compared with prior-based methods, our approach outperforms GCE-Pose~\cite{li2025gce} by 2.9\% on 5°2 \emph{cm} and 2.6\% on 5°5 \emph{cm}. And when compared with prior-free methods, MemPose surpasses the previous best method SpherePose~\cite{ren2025learning} by 1.7\% on 5°2 \emph{cm} and 2.1\% on 10°2 \emph{cm}, which also employs DINOv2~\cite{oquab2024dinov2} as the image backbone.
On the more challenging Housecat6D dataset, MemPose still outperforms the prior state-of-the-art methods on the precision of COPE. Specifically, when compared with previous best prior-based methods, MemPose surpasses GCE-Pose by 5.4\% on $IoU_{50}$ and 3.3\% on 10°2 \emph{cm}. And when compared with recent prior-free strong baselines, our method outperforms SpherePose by 3.8\% on 5°2 \emph{cm} and 11.7\% on 10°2 \emph{cm}, and surpasses AG-Pose~\cite{lin2024instance} by 4.6\% on $IoU_{50}$ and 3.4\% on $IoU_{75}$.
The significant improvements on these two benchmarks demonstrate the effectiveness of the proposed method.

\begin{table}[htbp] 
  \centering
  \begin{minipage}[c]{0.48\textwidth}
    \centering
    \setlength\tabcolsep{1pt}
    \scriptsize
    \begin{tabular}{l|c|ccccc}
        \toprule
         Methods  & $IoU_{75}$ & 5°2\emph{cm} & 5°5\emph{cm}  & 10°2\emph{cm} & 10°5\emph{cm}  \\
        \midrule
        SPD~\cite{tian2020shape}            & 20.3                 & 6.9        & 9.3       & 20.1     & 27.8     \\
        GPV-Pose~\cite{di2022gpv}       & -         & 14.1          & 21.5 & 23.8   & 41.1                                       \\
        SGPA~\cite{chen2021sgpa}            &56.4                &10.3         & 20.5        & 29.1   & 39.5  \\
        AG-Pose~\cite{lin2024instance}     & 36.2         &21.4         & 27.3       &29.1  & 40.1    \\
        Diff9D~\cite{liu2025diff9d}           &38.2               &25.5         & 30.5      & 32.5    & \underline{40.9}  \\
        MH6D~\cite{liu2024mh6d}           &\underline{41.9}                &\underline{27.0}         & \underline{31.2}       & \underline{34.4}   & 40.5   \\
        \midrule
        \textbf{MemPose}            & \textbf{46.2}    & \textbf{29.7}         & \textbf{33.5} & \textbf{35.8} & \textbf{43.2}  \\
        \bottomrule
    \end{tabular}
    \captionof{table}{\textbf{Comparisons with state-of-the-art methods on Wild6D dataset.} A higher value indicating better performance. Overall best results are in \textbf{bold} and the second best results are \underline{underlined}.}
    \label{tab:compare_sota_wild6d}
  \end{minipage}
  \hfill
  \begin{minipage}[c]{0.48\textwidth}
    \centering
    \includegraphics[width=\linewidth]{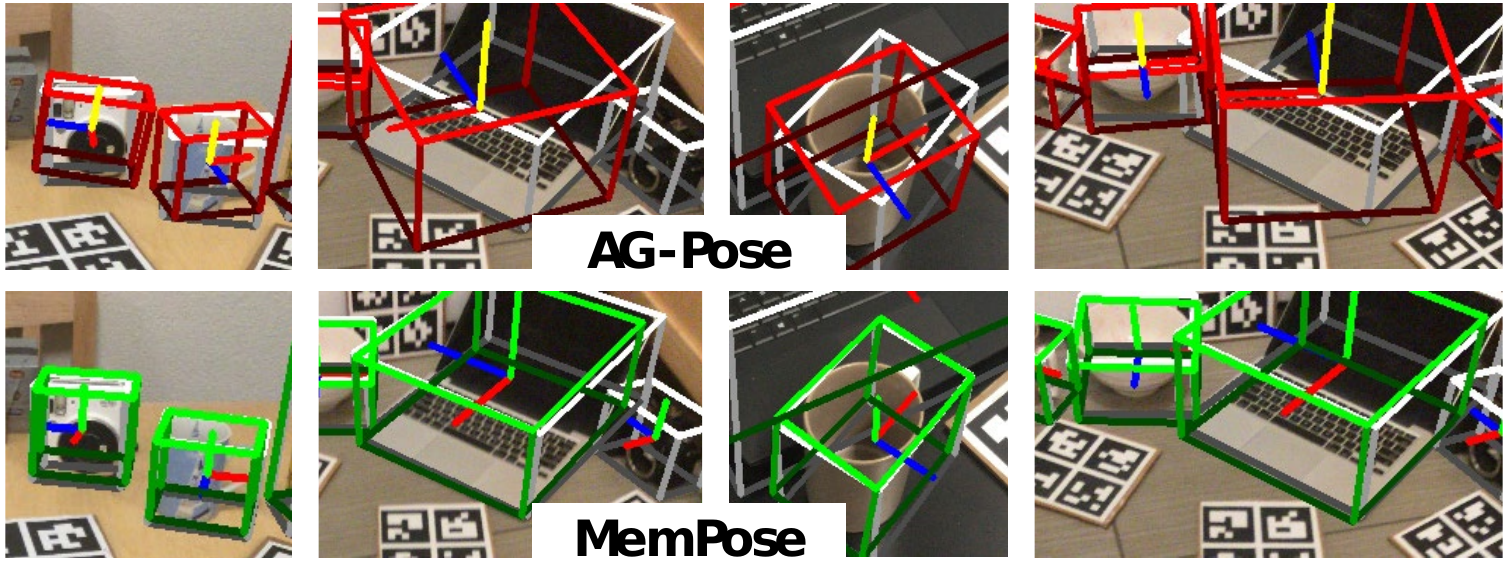}
    \captionof{figure}{Qualitative comparison on REAL275. We compare the predictions of MemPose and the baseline AG-Pose. The ground truth is marked by white borders.}
    \label{fig:vis}
  \end{minipage}
\end{table}

\begin{table}[htbp]
  \centering
  \label{tab:six_subtables}
  \begin{subtable}[t]{0.48\textwidth}
    \centering
    \setlength\tabcolsep{5pt} 
    \scriptsize
        \begin{tabular}{c|ccc}
            \toprule
            Type & $IoU_{75}$ & 5°2\emph{cm} & 5°5\emph{cm} \\
            \midrule
            \colorbox{mygray2}{Local} &\textbf{81.0} & \textbf{59.9}     &\textbf{67.7}        \\
            Global &80.5 & 58.0          &66.5         \\
            Fused &80.4 & 58.1    &66.8        \\
            \bottomrule
        \end{tabular}
    \caption{Effect of distinct stored memory.}
    \label{tab:ab_stored_memory}
  \end{subtable}
  \hfill 
  \begin{subtable}[t]{0.48\textwidth}
    \centering
    \setlength\tabcolsep{5pt}
    \scriptsize
        \begin{tabular}{c|ccc}
            \toprule
            Strategy & $IoU_{75}$ & 5°2\emph{cm} & 5°5\emph{cm} \\
            \midrule
            \colorbox{mygray2}{w/ warm-up} &\textbf{81.0} & \textbf{59.9}     &\textbf{67.7}        \\
            w/o warm-up &80.8 & 59.0          &66.0       \\
            Random & 80.4          &58.0    &65.0   \\
            \bottomrule
        \end{tabular}
        \caption{Effect of memory construction strategies.}
        \label{tab:ab_mem_construct}
  \end{subtable}
    \caption{Ablation studies on memory modules. Settings used in our final model are colored in \colorbox{mygray2}{gray}.}
\end{table}

\noindent
{\bf Results on CAMERA25 dataset.}
In \cref{tab:compare_sota_camera}, we compare our method with the existing approaches for category-level object pose estimation on CAMERA25~\cite{wang2019normalized} dataset. From the results, we can see that MemPose achieves the best performance. In detail, MemPose outperforms the current state-of-the-art method SpherePose by 2.2\% on 10°2\emph{cm}, and surpasses AG-Pose~\cite{lin2024instance} by 1.2\% on $IoU_{75}$, 1.8\% on 5°5\emph{cm} and 1.5\% on 10°2\emph{cm}, respectively.

\noindent
{\bf Results on Wild6D dataset.}
We further evaluate the proposed method on the Wild6D dataset, using CAMERA25 and REAL275 datasets for training.
\cref{tab:compare_sota_wild6d} reports the quantitative results of existing methods on Wild6D dataset. Once again, our MemPose achieves the best performance over state-of-the-art approaches by a large margin. Concretely, our MemPose exceeds previous best method MH6D~\cite{liu2024mh6d} by 4.3\% on $IoU_{75}$.
Moreover, when compared under the n°m \emph{cm} metric, our method achieved an average improvement of 2.5\% across all thresholds.
The superior performance on this more comprehensive and challenging real-world dataset further demonstrates the effectiveness of our approach.

\noindent
{\bf Qualitative Comparison.}
The qualitative results of AG-Pose and proposed MemPose are shown in \cref{fig:vis}. It can be observed that our method yields a more accurate pose estimation across diverse shapes and poses.
These exceptional outcomes further support the efficacy of our approach.

\subsection{Ablation Studies}
\label{sec:ablation_studies}
To further demonstrate the superiority of our method, we conduct comprehensive ablation studies on REAL275 dataset.

\begin{table}[htbp]
  \centering
  \label{tab:six_subtables}
  \begin{subtable}[t]{0.48\textwidth}
    \centering
    \setlength\tabcolsep{5pt}
    \scriptsize
        \begin{tabular}{c|ccc}
            \toprule
            Update Period & $IoU_{75}$ & 5°2\emph{cm} & 5°5\emph{cm} \\
            \midrule
            \colorbox{mygray2}{Pre-update}  & \textbf{81.0} & \textbf{59.9}     &\textbf{67.7}      \\
            Post-update  & 80.7          &59.0    &65.2   \\
            \bottomrule
        \end{tabular}
        \caption{Effect of different update periods.}
        \label{tab:ab_update_period}
  \end{subtable}
  \hfill 
  \begin{subtable}[t]{0.48\textwidth}
    \centering
    \setlength\tabcolsep{3pt}
    \scriptsize
        \begin{tabular}{cc|ccc}
            \toprule
            Method & Balance & $IoU_{75}$ & 5°2\emph{cm} & 5°5\emph{cm}\\
            \midrule
            FIFO & - & 80.7          &59.1    &66.0   \\
            \hline
            \multirow{2}{*}{Merge} & \colorbox{mygray2}{Noise} & \textbf{81.0} & \textbf{59.9}     &\textbf{67.7}       \\
            ~ & Top-$k$ & \textbf{81.0}          &59.2    &65.3   \\
            \bottomrule
        \end{tabular}
        \caption{Effect of update mechanism.}
        \label{tab:ab_update_mechanism}
  \end{subtable}

  \begin{subtable}[t]{0.48\textwidth}
    \centering
    \setlength\tabcolsep{5pt}
    \scriptsize
        \begin{tabular}{c|ccc}
            \toprule
            Size & $IoU_{75}$ & 5°2\emph{cm} & 5°5\emph{cm} \\
            \midrule
            \colorbox{mygray2}{All} & \textbf{81.0} & \textbf{59.9}     &\textbf{67.7}       \\
            Random-48 & 80.4          &58.0    &65.5   \\
            Random-16 & 80.1          &57.7    &65.0    \\
            Top-16 & 80.2   &57.9    &65.3 \\
            \bottomrule
        \end{tabular}
        \caption{Effect of retrieved size}
        \label{tab:ab_retrieval_size}
  \end{subtable}
  \hfill
  \begin{subtable}[t]{0.48\textwidth}
    \centering
    \setlength\tabcolsep{5pt}
    \scriptsize
        \begin{tabular}{c|ccc}
            \toprule
            Fusion & $IoU_{75}$ & 5°2\emph{cm} & 5°5\emph{cm} \\
            \midrule
            Add  & 80.4          &58.5    &65.3    \\
            \colorbox{mygray2}{Gate} & \textbf{81.0} & \textbf{59.9}     &\textbf{67.7}      \\
            Concat & 80.5         &58.0    &65.0    \\
            \bottomrule
        \end{tabular}
        \caption{Effect of memory fusion mechanism.}
        \label{tab:ab_fusion_mechanism}
  \end{subtable}
    \caption{Ablation studies on memory mechanisms. Settings used in our final model are colored in \colorbox{mygray2}{gray}.}
\end{table}

{\bf Effect of Distinct Stored Memory.}
We further present the different stored memory as mentioned in \cref{sec:memory_module}, the results are shown in \cref{tab:ab_stored_memory}. The table indicates that storing local features as memory entries achieves the best overall performance. This suggests that local features encode more explicit geometric information, which guides the model to focus on structural similarity rather than temporal proximity.

{\bf Effect of Memory Construction Strategies.}
As discussed in the previous section, the warm-up strategy refers to newly features are first fed into buffer until the buffer reaches full capacity, after which the update mechanism is activated.
\cref{tab:ab_mem_construct} reports the comparison of different memory construction strategies. The results show that the warm-up strategy can bring the best performance compared to the direct update strategy and random initialization. Consequently, the warm-up strategy provides a more reliable foundation for subsequent memory updates and retrieval.

{\bf Effect of Memory Update Mechanism.}
We then analyze the impact of the memory update stage. First, as shown in ~\ref{tab:ab_update_period}, performing memory updates before retrieval yields better performance than post-retrieval updates. This confirms that updating the memory in advance allows the stored representations to reflect the most informative and up-to-date geometric context, leading to more effective retrieval.
Second, we evaluate different memory update mechanisms, as reported in \cref{tab:ab_update_mechanism}. Compared to the FIFO strategy, the similarity-based token merge mechanism achieves superior performance on REAL275. 
To further prevent imbalanced updates like only a small subset of memory entries is repeatedly modified, we introduce lightweight balancing strategies. Specifically, the noise variant injects mild perturbations into cosine similarity computation, while the top-$k$ variant randomly selects an update target from the top-$k$ most similar entries. The noise-based token merge achieves the best overall results. 
These findings demonstrate that similarity-guided and balanced memory updates are crucial for maintaining a compact yet expressive memory buffer.

{\bf Effect of Retrieved Memory Size.}
We are also interested in how many memory entries of the buffer we need for retrieval. As shown in \cref{tab:ab_retrieval_size}, retrieving all memory entries yields the best performance, indicating that more memory entries allow the model to access richer geometric context. Although attention-guided selection (Top-16) outperforms random sampling with the same retrieval size, the performance gap with full retrieval remains.

\begin{table}[tb] 
  \centering
  \begin{minipage}[c]{0.48\textwidth}
    \centering
    \setlength\tabcolsep{2pt}
    \scriptsize
        \begin{tabular}{c|cccc}
            \toprule
            Buffer Length & $IoU_{75}$ & 5°2\emph{cm} & 5°5\emph{cm} & 10°2\emph{cm} \\
            \midrule
            0 (w/o mem) & 79.5 & 57.0          &64.6    &75.1     \\
            16 & 79.6 & 57.2    &64.9    &77.1    \\
            48 & 80.0 & 58.5         &66.5   &78.3    \\
            \colorbox{mygray2}{96} & \textbf{81.0} & \textbf{59.9}     &67.7    &\textbf{79.0}    \\
            256 & 80.8 &  59.1         &\textbf{67.9}    &78.6    \\
            \bottomrule
        \end{tabular}
        \caption{Effect of different buffer length. Setting 96 as the buffer length provides a balance between accuracy and efficiency.}
        \label{tab:ab_buffer_length}
  \end{minipage}
  \hfill
  \begin{minipage}[c]{0.48\textwidth}
    \centering
    \includegraphics[width=\linewidth]{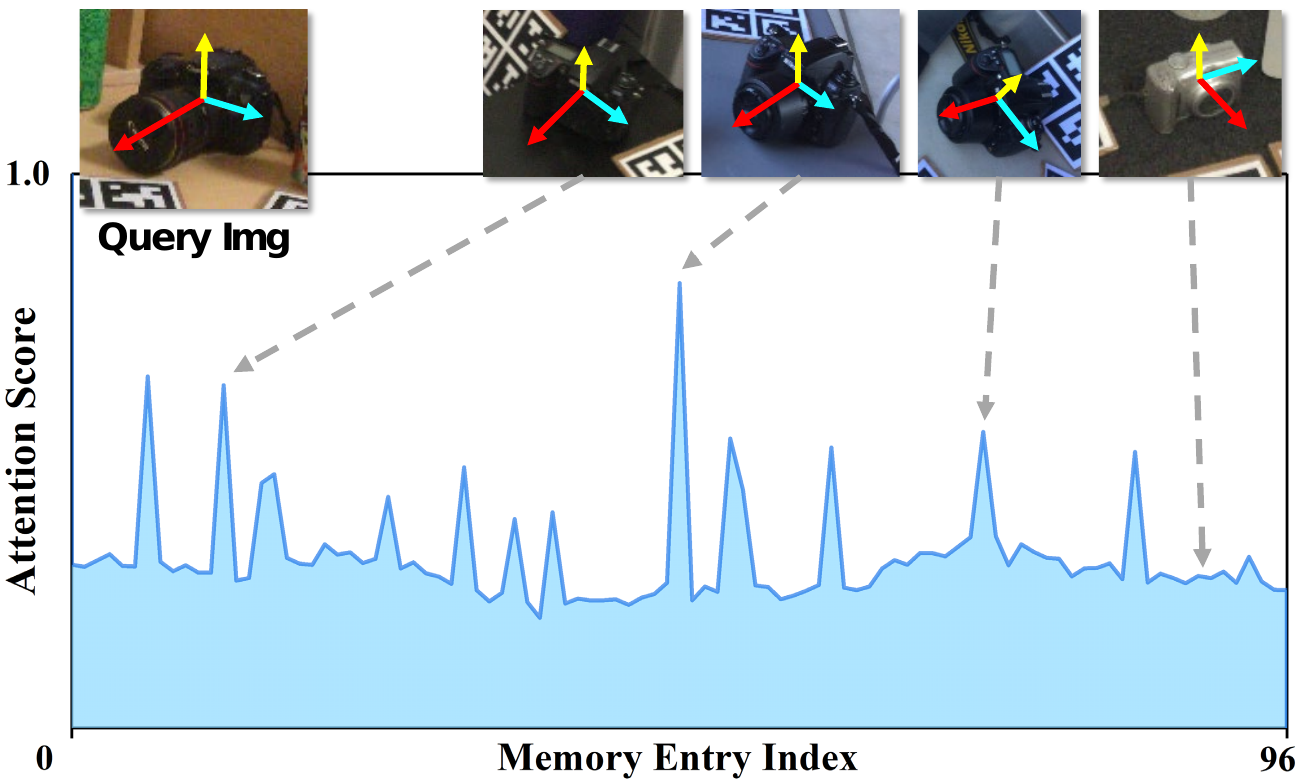}
    \captionof{figure}{Visualization of Memory Mechanism. The figure visualizes the retrieved memory elements and their attention weights during inference.}
    \label{fig:attn_score}
  \end{minipage}
\end{table}

{\bf Effect of Memory Fusion Mechanism.}
We present the ablation of different fusion mechanisms in \cref{tab:ab_fusion_mechanism}. From the results, we can observe that the proposed gating fusion achieves the best performance, outperforming the simple addition and direct concatenation.

{\bf Effect of Different Buffer Length.}
We investigate the impact of different buffer length and report the performance in \cref{tab:ab_buffer_length}. We adopt AG-Pose~\cite{lin2024instance} as baseline, which does not incorporate memory module, meaning the length is 0. Generally, we find that length 96 is good enough to train MemPose. Using longer length only results in comparable results but with much more computational overhead, \eg GPU memory and training time.
Therefore, we finally choose to use 96 as the buffer length, which provides a balance between accuracy and efficiency.

{\bf Discussion: Whether the memory mechanism retrieves relevant elements?}
To provide a direct view of how the memory mechanism functions, we present a qualitative case study that visualizes the retrieved memory elements and their attention scores during inference, as shown in \cref{fig:attn_score}.
The horizontal axis denotes the index of memory entries in the buffer, while the vertical axis represents the attention score assigned to each entry. During inference, the target object queries the memory buffer, and the attention scores reflect the relative importance of different memory entries for the current prediction.
From the visualization, we observe that memory entries with high attention scores correspond to instances whose poses and sizes are highly similar to those of the query object. In contrast, memory entries with low attention scores often exhibit significantly different or even opposite poses.
These observations suggest that the proposed memory mechanism effectively retrieves and leverages pose-relevant category-level geometric information, enabling the model to focus on the most informative instances.

\begin{table}[tb]
    \small
    \centering
    \setlength\tabcolsep{8pt}
    \begin{tabular}{c|l|c|c|c|c|c|c}
    \toprule
   ID & Method & Visual Enc. & Param.$\downarrow$ & $IoU_{75} \uparrow$ & 5°2\emph{cm}$\uparrow$ & TT$\downarrow$  & IT$\uparrow$\\
    \midrule
    1        &AG-Pose      &  \multirow{3}{*}{\doubleline{DINOv2}{(VIT-S/14)}}     &223M     &80.1         &57.0 &47.5 &\textbf{35}   \\
    2         &\colorbox{mygray2}{ours}     &~        & 225M     &\textbf{81.0}       &	\textbf{59.9} &47.8 &\underline{33}   \\
    3        &ours$^{\dag}$      &  ~      & 225M     &\underline{80.5}         &\underline{57.8} &47.8 &\underline{33}   \\
    \midrule
    4         & AG-Pose        & \multirow{2}{*}{ResNet18}      & \textbf{220M}      &77.6       &56.2 &\textbf{46.9} &\textbf{35}   \\
    5         & ours        & ~      & \underline{222M}      &80.3      &57.4 &\underline{47.1} &\underline{33}   \\
    \bottomrule
    \end{tabular}
    \caption{Detailed comparison results on REAL275.  ‘$\dag$’ represents replacement of memory module with MLPs of the same number of parameters. TT: Traning Time (min/epoch), IT: Inference Speed (frame/sec). Overall best results are in \textbf{bold} and the second best results are \underline{underlined}. Default settings are colored in \colorbox{mygray2}{gray}.
    }
    \label{tab:detail_comparison}
\end{table}

{\bf Detailed Comparison with AG-Pose.}
Following common practice in this domain, we report the standard accuracy metrics in \cref{tab:compare_sota}. To provide a more transparent and comprehensive comparison beyond accuracy, \cref{tab:detail_comparison} further summarizes key practical factors, including the visual encoder type, the total number of parameters, training time per epoch (TT), and inference throughput (IT, FPS).
As confirmed in (\#1) and (\#2) of \cref{tab:detail_comparison}, the memory module introduce only negligible parameter overhead when using the same DINOv2 ViT-S/14 backbone (225M \vs 223M), while the running time remains nearly unchanged (33 \vs 35 in FPS). 
More importantly, this gain does not come at the cost of efficiency: our method achieves comparable training time (47.8 vs. 47.5 min/epoch) while maintaining essentially the same inference budget and even slightly higher throughput (33 vs. 35 FPS).
Notably, the memory buffer stores non-parametric entries and therefore does not introduce additional learnable weights; the parameter change mainly comes from lightweight integration components, making the overhead minimal in practice.

To ensure a fair comparison, we also replace the memory module with MLPs that have the same number of parameters (\#3). These results indicate that the gains stem from the proposed memory mechanism rather than from additional parameters.
What's more, we include additional results with ResNet18. Specifically, our method still outperforms AG-Pose with ResNet18 setting (\#4 \vs \#5), further supporting the efficacy of our approach.

{\bf Memory in Inference.}
During training, the category-specific memory buffer is registered as a non-trainable buffer and updated online with incoming samples. Once the corresponding category buffer is full, memory update and retrieval are activated to provide additional geometric context for pose estimation.
In principle, the same memory mechanism can also be applied during inference. As new objects are perceived, the memory buffer can be continuously updated, allowing the model to accumulate additional category-level geometric knowledge from the test stream.
However, to ensure a fair comparison with previous methods, all main results reported in this paper are obtained under a frozen-memory inference protocol, where the memory buffer is fixed after training and is not updated on the test set. 
\begin{table}[htbp]  
    \centering
    \setlength\tabcolsep{4pt}
    \begin{tabular}{c|cccc}
        \toprule
        Method & $IoU_{75}$ & 5°2\emph{cm} & 5°5\emph{cm} & 10°2\emph{cm}\\
        \midrule
        AG-Pose~\cite{lin2024instance} &80.1       & 57.0         &64.6         & 75.1      \\
        \colorbox{mygray2}{MemPose (frozen memory)} &81.0 & 59.9          &67.7    &79.0    \\
        MemPose (updating memory) & \textbf{81.2}          &\textbf{63.6}    &\textbf{68.3} &\textbf{80.9}   \\
        \bottomrule
    \end{tabular}
    \caption{Performance of updating memory buffer during inference on REAL275. Settings used in our final model are colored in \colorbox{mygray2}{gray}.}
    \label{tab:performance_inference}
\end{table}
Therefore, our standard evaluation does not introduce test-set leakage or test-time adaptation effects. Additional experiments and detailed analysis of inference-time memory updating are provided in \cref{tab:performance_inference}.

\section{Conclusion}
In this work, we perform a comprehensive analysis of existing parametric COPE paradigms and introduce a core essential: the dynamic mechanism to accumulate and reuse category-level geometric memory, which is essential for robust COPE.
This analysis reveals that relying solely on static shape priors or fixed model parameters limits the ability of current approaches to generalize across diverse object instances.
Based on these insights, we propose MemPose, a simple yet effective memory-augmented framework that explicitly incorporates category-level geometric memory into the COPE pipeline.
MemPose provides meaningful contextual support for pose estimation, enabling the model to leverage accumulated geometric knowledge from previous observations.
Extensive experimental results on three challenging benchmarks demonstrate the effectiveness of the proposed method.


\section*{Acknowledgements}
This paper is supported by the National Natural Science Foundation of China (No.62233013, 62333017, 62403358).

%
%
\bibliographystyle{splncs04}
\bibliography{main}
\end{document}